\documentclass[runningheads]{llncs}
\usepackage{graphicx}
\begin{document}
\title{Evaluating Table Structure Recognition: A New Perspective}
\author{Tarun Kumar \and Himanshu Sharad Bhatt}
\authorrunning{T. Kumar et al.}
\institute{American Express AI Labs, India\\
\email{\{tarun.kumar16, himanshu.s.bhatt\}@aexp.com}}
\maketitle              % typeset the header of the contribution
\begin{abstract}
Existing metrics used to evaluate table structure recognition algorithms have shortcomings with regard to capturing text and empty cells alignment. In this paper, we build on prior work and propose a new metric - TEDS based IOU similarity (TEDS (IOU)) for table structure recognition which uses bounding boxes instead of text while simultaneously being robust against the above disadvantages. We demonstrate the effectiveness of our metric against previous metrics through various examples.

\end{abstract}
\section{Introduction}
A huge amount of information flows through enterprise documents; thus, it is imperative to develop efficient information extraction techniques to extract and use this information productively. While documents comprise multiple components such as text, tables, figures etc.; tables are the most commonly used structural representation that organize the information into rows and columns. It captures structural and geometrical relationships between different elements and attributes in the data. Moreover, important facts/numbers are often presented in tables instead of verbose paragraphs. For instance, tables in financial domain are a good example where different financial metrics such as ``revenue", ``income" etc. are presented for different quarters/years. Extracting the content of a table into a structured format (csv or JSON) \cite{gao2019icdar}, \cite{gobel2013icdar}, \cite{jimeno2021icdar} is a key step in many information extraction pipelines. 

Unlike traditional machine learning problems where the output is a class (classification) or number (regression), the outcome of a table parsing algorithm is always a structure. There needs to be a way to compare one structure against another structure and define some measure of ``similarity/distance" to evaluate different methods. A number of metrics quantifying this ``distance" have been proposed in literature and multiple competitions. Existing metrics evaluates the performance of table parsing algorithms using the structural and textual information. This paper presents the limitation of existing metrics based on their dependence on the textual information. We emphasize that textual information introduces additional dependency on the OCR (text detection/recognition), which is a separate area in itself and should not be included in evaluating how good is the detected table structure. This paper presents a ``true" metric which is agnostic to the textual details and accounts only for the layout of cells in terms of its row number/column number and bounding box. 

\vspace{-6pt}
\section{Existing Metrics in Table Parsing}
Two of the existing metrics are adjacency relation set-based F1 scores with different definitions of the set. They break and linearize the table structure into two dimensions, one along the row and one along the column. Adjacency Relation (Text) \cite{gobel2013icdar} computes pair-wise relations between non-empty adjacent cells and the relation is considered correct only if the direction (horizontal/vertical) and text of both the participating cells match. It does not take into account empty cells and multi-hop cell alignment. Adjacency Relation (IOU) \cite{gao2019icdar} is a text-independent metric where original non-empty cells are mapped to predicted cells by leveraging (multiple) IOU thresholds and then adjacency relations are calculated. This metric takes a weighted average of the computed F1-scores at different IOU thresholds \{0.6, 0.7, 0.8, 0.9\}. Finally, the predicted relations are compared to the ground truth relations and precision/recall/F1 scores are computed.

The third metric considers the structure as a HTML encoding of the table. In this representation, the table is viewed as a tree with the rows being the children of the root $<table>$ node, and cells being the children (represented by $<td>[text]</td>$) of the individual rows. A Tree edit-distance (TEDS) metric \cite{zhong2020image} is proposed which compares  two trees and reports a single number summarizing the similarity.

While there are other metrics used in literature such as BLEU-4 \cite{li2019tablebank} (which is more language based), this paper only considers the above three most widely used metrics for evaluating the performance of table structure recognition.

\section{Proposed Metric}
This paper highlights the limitations of the previous metrics and also proposes a new metric, Tree-Edit-Distance Based Similarity with IOU \textit{(TEDS-IOU)}, for evaluating table structure recognition algorithms. The paper also demonstrates how \textit{TEDS-IOU} addresses the limitations of existing metrics. 

\begin{figure}[t]
\begin{center}
  \includegraphics[width=0.5\linewidth]{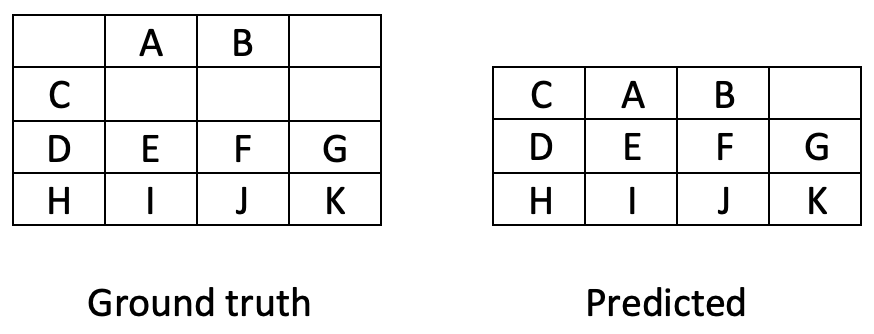}
   \vspace{-10pt}
  \caption{Original table and an example prediction for the same. For Adjacency relation (Text), the characters can be considered as representing the text inside cells. For Adjacency relation (IOU), characters can be considered as labels representing cells.}
  \label{fig:digitalization}
  \vspace{-16pt}
 \end{center}
\end{figure}

Table \ref{tab1} describes the limitations of the commonly used metrics in table structure recognition literature. For example, in figure \ref{fig:digitalization}, even though the predicted table missed one entire row and $4$ empty cells, in terms of adjacency relations, the only extra relation in the predicted table is the \{C, A, Horizontal\}, where `Horizontal' is the direction of relation. This only affects precision but the recall is still $100$\% which clearly should have been penalised. Also, in the case of the IOU based metric, lets assume label mapping, i.e. cell represented by ``C" in ground truth is a mapped to the ``C" cell in predicted table using IOU thresholds. We still have that same extra relation \{``C", ``A", Horizontal\}, where `Horizontal' is the direction, which demonstrates the inability to capture empty cells and mis-alignments. We should note that metric is still better then the text-based version, since it does not rely on comparing text. Accurately detecting and recognizing text (OCR) is a separate field in itself, while in table structure recognition, we are primarily interested in localizing the cell boundaries and assign text to them.

\begin{table}[t]
\caption{ Existing metrics in literature and their limitations}\label{tab1}
\begin{tabular}{p{0.32\textwidth}|p{0.66\textwidth}}
\hline
\textbf{Metric} & \textbf{Limitations}\\
\hline
{Adjacency Relation (Text)} & {Doesn't handle empty cells, misalignment of cells beyond immediate neighbours \& text dependent}\\
{Adjacency Relation (IOU)} & {Doesn't handle empty cells, misalignment of cells beyond immediate neighbours}\\
{TEDS (Text)} & {Text dependent but less strict due to Levenshtein distance}\\
\hline
\end{tabular}
 \vspace{-16pt}
\end{table}
TEDS (Text) metric solved the shortcomings of previous metrics with regard to empty cells and multi-hop mis-alignments \cite{zhong2020image}. In TEDS, all cells, with or without text are considered, thereby also including empty cells as part of computation. So, TEDS (text) will penalise the absence of a row and all the alignment mismatches when comparing ground truth table against predicted table in figure \ref{fig:digitalization}. But it computes the edit distance between cells' texts as compared to the exact match in Adjacency Relation (Text).

Table structure recognition algorithms aim at predicting the location (bounding boxes) of cells and their logical relation with one another, irrespective of the text in the cell. Therefore, the evaluation metric should not penalize an algorithm for inaccuracies in text. With this observation, this paper propose TEDS (IOU) which replaces the string edit distance between cells' text with the IOU distance between their bounding boxes. This effectively, removes dependency on text or OCR, while also preserving the benefits of the original TEDS (text) metric. Specifically, we compute TEDS (IOU) as follows: cost of insertion \& deletion operations is 1 unit; while substituting a node $n_s$ with $n_t$ - cost of edit is 1 unit if either $n_s$ or $n_t$ is not $<td>$, cost of edit is 1 unit if both $n_s$ \& $n_t$ is $<td>$ and the column span or row span of $n_s$ \& $n_t$ is different, otherwise, cost of edit is $1 - IOU(n_s.bbox,  n_t.bbox)$. Finally,
  \begin{equation}
  TEDS\_IOU(T_a, T_b) = 1 - \frac{EditDistIOU(T_a, T_b)}{max(|T_a|, |T_b|)}
  \end{equation}
TEDS (IOU) $\in [0,1]$, the higher the better. $|.|$ denotes cardinality. IOU distance $(IOU_d = 1 - IOU)$ being a Jaccard index \cite{kosub2019note}, is a metric as it satisfies:
\begin{enumerate}
\item $IOU_d(A, B) = 0 \iff A = B  \qquad \qquad \qquad \qquad \qquad \; \;$ $Identity$
\item $IOU_d(A, B) = IOU_d(B, A) \qquad  \qquad \qquad \qquad \qquad \; \; \; \; \; \;$ $Symmetry$
\item $IOU_d(A, C) <= IOU_d(A, B) + IOU_d(B, C) \qquad \qquad \; \,$ $Triangle\;Inequality$
\end{enumerate}

\begin{figure}[t]
\begin{center}
  \includegraphics[width=0.8\linewidth]{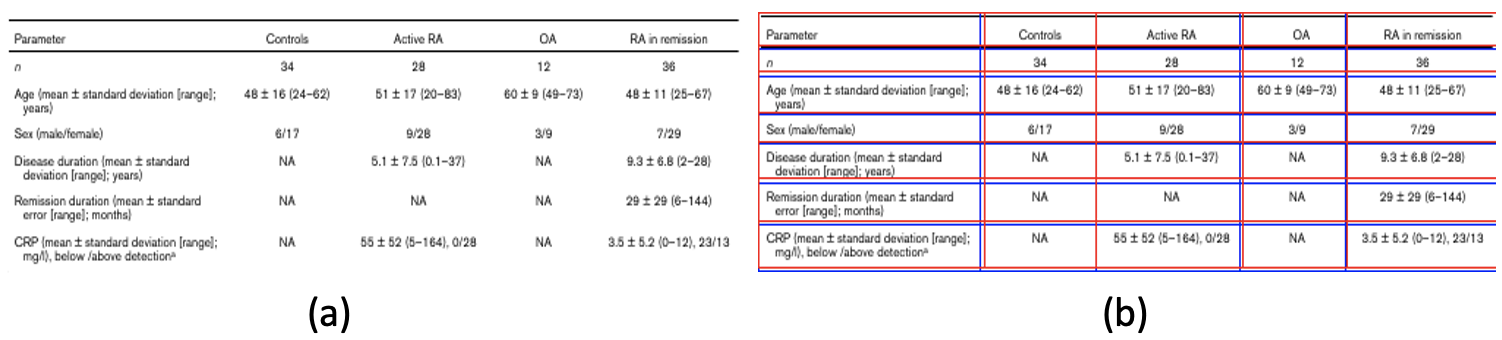}
   \vspace{-10pt}
  \caption{(a) is a table from PubTabNet dataset. In (b), red lines denote the predicted structure and blue lines depict the true structure.}
  \label{test}
  \vspace{-16pt}
 \end{center}
\end{figure}

To demonstrate the effectiveness of the proposed TEDS (IOU) metric, we compute the all four metrics for the predicted table in figure \ref{test}(b). In the example above, we had known OCR issues where it was unable to recognize the $\pm$ symbol (it got recognized as +) and all the cells with``NA" were detected as empty. Adjacency Relation (Text) got a very poor score of $13.7$ F1 due to the exact text match constraint. Adjacency Relation (IOU), being text independent, is more robust and achieves a Weighted Avg. F1 of $59.8$. TEDS (text) matches text through edit distances, therefore, for it, only the ``NA" cells gave high edit distance (of 1) and it scores $71.6$ on this table. TEDS (IOU) being text independent and computing the IOU distance between cells, assigns a higher score of $80.6$ which seems to be the most representative one of the prediction.

\section{Discussion \& Future Work}

We proposed a new metric for table structure recognition and demonstrated its benefits against existing metrics. As future steps, we plan to compare these metrics across different datasets and models. A possible extension of this work can be to introduce different thresholds for the IOU as in Adjacency Relation (IOU), instead of using absolute numbers.
\vspace{-10pt}

\bibliographystyle{splncs04}

\end{document}